\documentclass[conference]{IEEEtran}

\usepackage{comment}
\usepackage{adjustbox}
\usepackage{url}
\usepackage{makecell}
\usepackage{xcolor}
\usepackage{textcomp}
\usepackage{siunitx}
\usepackage{graphics}
\usepackage{multirow}
\usepackage{booktabs}
\usepackage{lscape}
\usepackage{graphicx}
\usepackage{verbatim}
\usepackage[hidelinks]{hyperref}

\usepackage{amssymb}

\usepackage{tikz}

\usepackage[flushleft]{threeparttable}
\usepackage{multirow}
\usepackage[normalem]{ulem}
\useunder{\uline}{\ul}{}
\usepackage{subfig}
\usepackage{amsmath,amssymb,amsfonts}
\usepackage{algorithmic}
\usepackage{graphicx}
\usepackage{pgfplotstable}
\usepackage{textcomp}
\usepackage{hyperref}
\usepackage[backend=bibtex, minnames=1, maxnames=3, doi=false, isbn=false, style=ieee]{biblatex}

\usepackage{geometry}
\ifCLASSINFOpdf
\else
\fi
\begin{document}
% \title{A Novel Method to Measure Students' Lecture Understanding Using Features from EEG and NLP}

% \title{A Novel Method to Measure Students' Lecture Comprehension Using EEG and NLP Features}

\title{Combining EEG and NLP Features for Predicting Students' Lecture Comprehension using Ensemble Classification}

% \author{\IEEEauthorblockN{Phantharach~Natnithikarat}
% \IEEEauthorblockA{Ruamrudee International School\\
% Bangkok, Thailand\\
% }
% \and
% \IEEEauthorblockN{Sanurak~Natnithikarat}
% \IEEEauthorblockA{University of California, Berkeley\\
% California, USA\\
% }
% \and
% \IEEEauthorblockN{Theerawit Wilaiprasitporn}
% \IEEEauthorblockA{Vidyasirimedhi Institute of Science \\ and Technology\\
% Rayong, Thailand\\
% }
% \and
% \IEEEauthorblockN{Supavit Kongwudhikunakorn}
% \IEEEauthorblockA{Vidyasirimedhi Institute of Science \\ and Technology\\
% Rayong, Thailand\\}}



% Remove author names as it is a double-blind review.

\author{\IEEEauthorblockN{Phantharach~Natnithikarat,
Theerawit~Wilaiprasitporn\IEEEauthorrefmark{2} and 
Supavit~Kongwudhikunakorn\IEEEauthorrefmark{2} }
\IEEEauthorblockA{Ruamrudee International School, Bangkok, Thailand}
\IEEEauthorblockA{\IEEEauthorrefmark{2}Vidyasirimedhi Institute of Science and Technology, Rayong, Thailand} 
}

% used for special paper notices

\maketitle

\begin{abstract}
% new v.
Electroencephalography (EEG) and Natural Language Processing (NLP) can be applied for education to measure students' comprehension in classroom lectures; currently, the two measures have been used separately.
In this work, we propose a classification framework for predicting students' lecture comprehension in two tasks:
(i) students' confusion after listening to the simulated lecture and (ii) the correctness of students' responses to the post-lecture assessment.
The proposed framework includes EEG and NLP feature extraction, processing, and classification.
EEG and NLP features are extracted to construct integrated features obtained from recorded EEG signals and sentence-level syntactic analysis, which provide information about specific biomarkers and sentence structures.
An ensemble stacking classification method -- a combination of multiple individual models that produces an enhanced predictive model -- is studied to learn from the features to make predictions accurately.
Furthermore, we also utilized subjective confusion ratings as another integrated feature to enhance classification performance.
By doing so, experiment results show that this framework performs better than the baselines, which achieved F1 up to 0.65 for predicting confusion and 0.78 for predicting correctness, highlighting that utilizing this has helped improve the classification performance.

% Old v.
% Electroencephalography (EEG) and Natural Language Processing (NLP) are tools often applied for educational purposes to test students’ understanding in classroom environments and analyze linguistic information.
% EEG and NLP have been used separately to provide biological evidence and quantitative measurement.
% In order to integrate the characteristics of a biomarker specific to individuals and the analysis of words, this work aims to utilize both signals from the brain and features of language to predict students’ confusion after listening to separate sentences in a simulated lecture and find patterns within sentence structures that allow for optimum understanding.
% The results from the experiment show that it is indeed possible to identify confusion by using EEG as biomarkers and that including NLP features improves model performance.
% The model used achieved accuracies of 85\% and 85\% for predicting confusion and the correctness of students’ answers, which is the highest on similar tasks. 
\end{abstract}

\begin{IEEEkeywords}
Electroencephalograph (EEG), Natural Language Processing (NLP), Brain-Computer Interface (BCI), Lecture Comprehension, Classroom Lecture, Ensemble Classification
\end{IEEEkeywords}

% \fi
\IEEEpeerreviewmaketitle

\section{Introduction}
As the transfer of verbal information from educators to students forms a core aspect of education, this study aims to propose a classification framework to enhance this process. There are currently two metrics used to measure students' lecture comprehension \cite{gennlp, geneeg} electroencephalography (EEG) and natural language processing (NLP).

EEG is a technique used to measure the electrical activity of the brain \cite{consumer_eeg}. 
It uses electrodes placed on or below the scalp to record activity with a high temporal resolution, allowing researchers to detect changes during specific time periods quickly \cite{knot_tim}. 
Signals from EEGs can be used to measure the level of students' engagement and attention in classroom settings \cite{Ko2017} by monitoring changes in the EEG spectra \cite{Wang2021}. 
% Recent research on EEG and student confusion has achieved an accuracy of 75\% in predicting confusion \cite{mooc}. 
% The dataset included ratings of confusion on a scale of 1-7 after each video session. 

Another tool used is NLP, which helps computers understand, manipulate, and interpret human language. 
Some researchers have explored the application of NLP in educational settings, highlighting its effectiveness in addressing various challenges and improving learning, writing, analysis, and assessment processes \textcolor{black}{\cite{nlpeducation}}.
The study emphasizes the integration of NLP in different educational contexts, including research, science, linguistics, e-learning, and evaluation systems \cite{Alhawiti2014}.
NLP can be used to determine the structure of sentences, the different parts of speech, and sentimental analysis to determine words' positive and negative connotations \cite{Jia2020}.

Current methods of separately measuring students' understanding in classroom lectures with NLP and EEG have limitations since the former provides a quantitative measurement but lacks biological evidence specific to individuals, while the latter is a biomarker for specific individuals but needs to reflect the actual data delivered. 
Thus, to improve students' understanding of content and teachers' method of delivery, features extracted from both NLP and EEG could be simultaneously used to determine the information exchange along with time-specific data for comparing the structure of teachers' lectures to students' brain responses.

This work proposes a classification framework that combines EEG and NLP features to determine students' confusion levels after listening to simulated classroom lectures and predict correctness of students' response to the post-lecture assessment.
To compare against the predictions, the subjective confusion ratings and assessment responses from students are also collected.
Furthermore, to enhance classification performance in predicting correctness, the self-rated confusion levels are also combined with the previously mentioned features.
In addition, an ensemble stacking model is specifically utilized to learn the integrated features and perform the classification task.
To evaluate the effectiveness of the proposed method, we have performed experiments against several baseline methods in which the results show that this study outperforms all other baselines.

The following sections discuss the materials and methods of this study, followed by the experimental results, discussion, and conclusion presented as the final section.

% The contribution of this work is summarized as follows:
% \begin{itemize}
%     \item The provision of an innovative approach that combines EEG biomarkers with NLP to assess student comprehension.
%     \item The integration of subjective self-evaluations with objective correctness measures, creating a real-time metric for comprehension.
%     \item The potential to reshape traditional educational paradigms through detailed, sentence-by-sentence evaluations.
%     \item The establishment of a foundation for subsequent investigations in this domain.

% \end{itemize}
\begin{figure}
    \centering
    \includegraphics[width= \columnwidth]{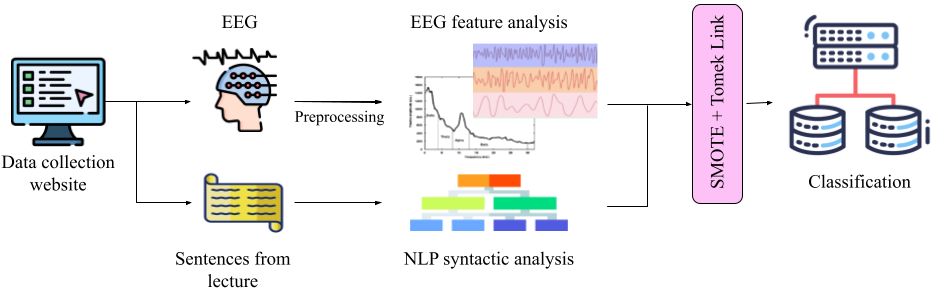}
    \caption{Overview of this study.}
    \label{fig:work_overview}
\end{figure}

\section{Materials and Method}

This study aims to measure students' lecture comprehension in a simulated classroom setting by using the proposed classification framework to learn from the integrated features of both the recorded EEG signals and the complexity of the lecture sentence structures.
There are two tasks in this study -- (i) to determine the correctness of post-lecture assessment responses and (ii) to determine the level of confusion.
The overview of this study is presented in \autoref{fig:work_overview}.

\subsection{Participants}
There are 21 participants (ages 16-21; 10 males and 11 females) recruited from international schools across Bangkok for EEG data collection during audio sessions. Before the experiment, participants were required to fill out a questionnaire assessing their English proficiency. This was done to ensure they had a comparable level of understanding, thereby reducing potential confounding variables.

Furthermore, to bolster the dataset specifically for NLP analysis, a separate group of 27 participants (ages 15-20; 12 males and 15 females) was engaged. These participants were provided with a website to rate and test their comprehension of different sentences. The rationale behind this approach was that these individuals could not physically participate in EEG data collection. By incorporating their input, we aimed to enrich the data pool for NLP, ensuring a more robust and comprehensive linguistic analysis.

This study was reviewed and approved by Ruamrudee International School.
All participants provided informed consent in accordance with the Declaration of Helsinki.
Before the experiment, the experimenter explained instructions and protocol to all participants.

% Example of how to add figure
% \begin{figure}
%     \centering
%     \includegraphics[width=0.5 \columnwidth]{Fig/cat.jpg}
%     \caption{Cat Figure}
%     \label{fig:cat_fig}
% \end{figure}

% The image of cat is shown in Figure~\ref{fig:cat_fig}.

% \begin{table}[t]
% \centering
% \caption{Summary of task characteristics.}
% {\begin{tabular}{lcc}
% \toprule[0.2em] 
% \multicolumn{1}{c}{\textbf{Task}} & \multicolumn{1}{c}{\textbf{Number of Trials}} & \multicolumn{1}{c}{\textbf{Dataset Dimension}} \\  \midrule[0.1em]
% Eyes-Closed & 25 & Set\#1: $54\times25\times30\times1,280$ \\
% Eyes-Opened & 25 & Set\#1: $54\times25\times30\times1,280$ \\
% Mental Imagery & 27 & Set\#1: $54\times27\times30\times1,280$ \\
%  \bottomrule[0.2em]
% \end{tabular}}
% \label{tab:subject}
% \end{table}

% The table of subject data is shown in Table~\ref{tab:subject}.

% EEG was mentioned in paper \cite{eeg_knot}.

% \subsubsection{aaa}

% \begin{itemize}
%     \item one
%     \item two
% \end{itemize}

\subsection{Data Acquisition}
In this study, the MUSE EEG headband (a consumer-grade EEG measuring sensor proved to be useful as a research tool \cite{consumer_eeg}) was used to record participants' EEG while performing the experiment. 
Four EEG channels (AF7, AF8, TP9, TP10) were collected at the sampling rate of 256 Hz, with FPz as a reference channel.
The experiments were conducted in an isolated room while participants were seated in a comfortable armchair and instructed to avoid unnecessary movements.

A data collection website was designed and developed specifically for this experiment to collect participants' subjective ratings and comprehension scores.
There were two separate websites used. The primary website was engineered with an integrated architecture, enabling simultaneous EEG data capture and administration of listening comprehension tasks. This design ensured the synchronous recording of neural activity, providing a comprehensive temporal alignment with the auditory stimuli presented. Conversely, the secondary website was singularly dedicated to the execution of listening comprehension tasks. This bifurcation was strategically implemented to cater to subsets of participants, ensuring that those not subjected to EEG instrumentation experienced the auditory stimuli in a controlled yet uninfluenced environment.

\begin{figure}
    \centering
    \includegraphics[width=1 \columnwidth]{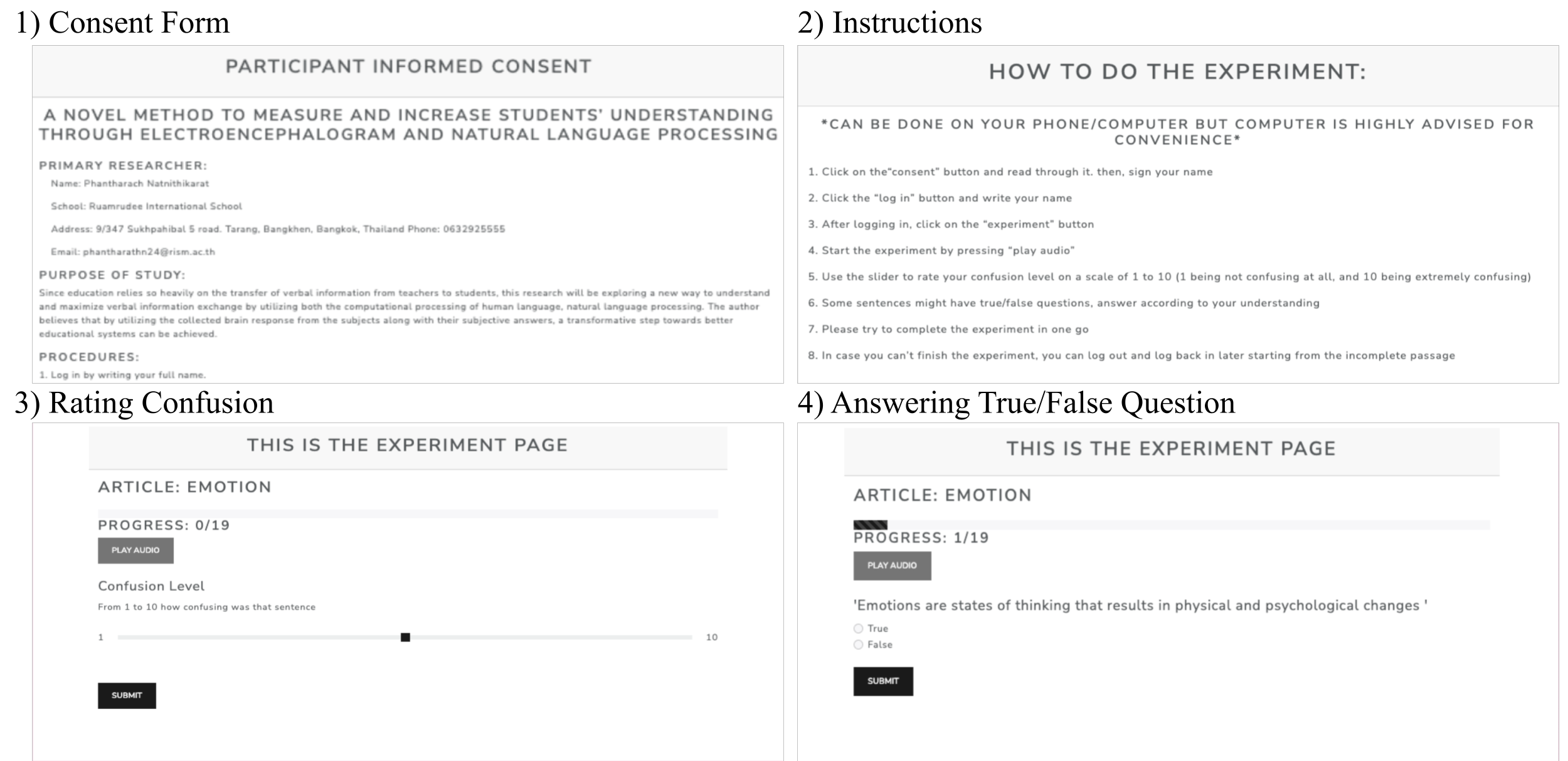}
    \caption{Website user interface and a sample question.}
    \label{fig:website}
\end{figure}

The website user interface is shown in Figure~\ref{fig:website}. 
The website helps facilitate the experiment, simulate the teaching environment, and collect data regarding the self-rated confusion levels and post-lecture assessment responses.

\color{black}
To synchronize the EEG data, we used the Open Sound Control protocol to align audio files with the EEG signals obtained from MUSE. These texts were converted into auditory stimuli using the FastSpeech2 algorithm \cite{fastspeech}. This algorithm allows us to pinpoint the timestamps of the word in each sentence.

For minimal latency, the experiment webpage runs locally during EEG recordings. By noting the timestamp when the user initiates playback and the audio track's duration, we can synchronize the audio data with the EEG stream.

\color{black}

\begin{figure}
    \centering
    \includegraphics[width=1 \columnwidth]{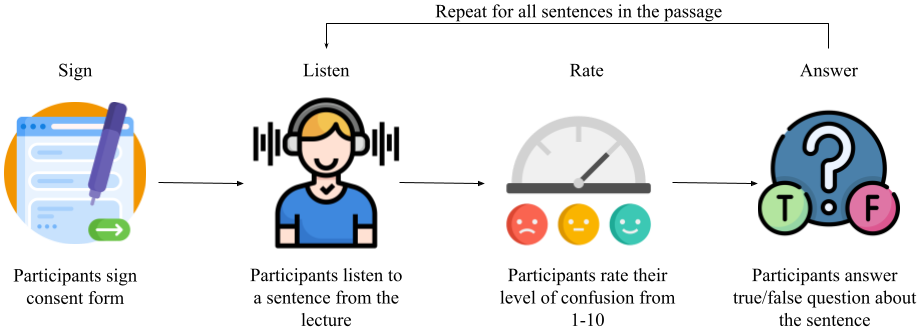}
    \caption{Experimental protocol of this study.}
    \label{fig:experiment_protocol}
\end{figure}

\subsection{Experimental Protocol}

There were four primary stages in the experimental protocol. Participants first provided their formal consent by signing the requisite participation agreement. Then, they were exposed to segments of a lecture, specifically on a sentence-by-sentence basis rather than the lecture in its entirety. After each sentence, participants were prompted to evaluate their level of comprehension ambiguity using a scale ranging from 1 to 10. On this scale, a score of 1 signified minimal confusion, whereas a score of 10 indicated maximal confusion. This quantification was facilitated through a range slider, ensuring precision in the participants' responses. Finally, participants responded to a binary (true/false) question on the content of the immediately preceding sentence.
The experimental protocol of this study is shown in \autoref{fig:experiment_protocol}.

The experimental procedure of this study took a total duration of 30 minutes.
This time frame was systematically divided into 5 minutes for meticulous setup and calibration of the EEG apparatus and 25 minutes for performing the experiment.
During the experiment, the participants were asked to listen to the set of passages, each lasting for 5 minutes, during which electroencephalogram (EEG) data was continuously recorded.

Each participant was presented with a series of five distinct passages. This collection comprised two standardized passages that were commonly administered across all participants, complemented by three passages that were randomized to introduce variability in the auditory stimuli. This design aimed to ensure a balance between uniformity in experimental conditions and the introduction of diverse auditory stimuli.

The textual content of these passages was sourced from the Stanford Question Answering Dataset (SQuAD) \cite{squad}, a renowned reading comprehension dataset.  On average, each passage consisted of 9-10 sentences.

\begin{figure}
    \centering
    \includegraphics[width=0.75 \columnwidth]{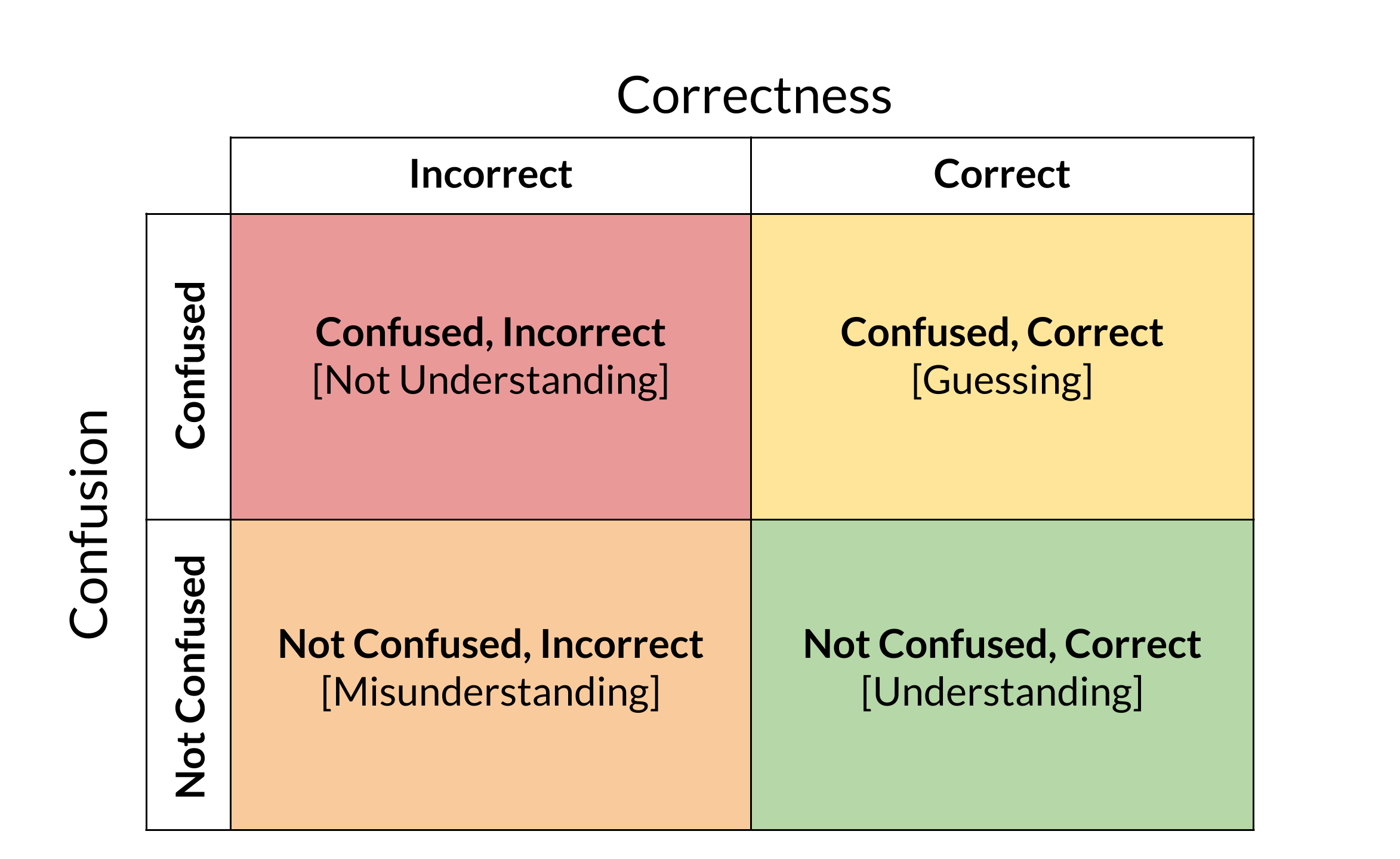}
    \caption{Adaptation of the known-unknown matrix to interpret understanding.}
    \label{fig:knownunknown}
\end{figure}

In this work, we employed a modified version of the well-known known-unknown matrix, originally introduced by \cite{Zero} to illustrate an individual's knowledge on a particular field, as depicted in \autoref{fig:knownunknown}, served as the basis for this adaptation.
We used the recorded students' self-rated confusion levels and correctness of post-lecture assessment responses to determine their comprehension using this matrix.
The evaluation from this matrix helped determine whether the students had a clear understanding, were confused, or simply guessed the correct answer.

\subsection{Data Preprocessing}
We initially normalized the EEG signal and applied a Butterworth bandpass filter ranging from 4Hz to 80Hz. In addition, we removed electrooculographic (EOG) artifacts caused by eye movements and blinks to improve the quality of the EEG data.

\subsection{Feature Extraction}
\subsubsection{EEG Features}

Spectral density estimation remains a fundamental aspect of EEG feature extraction. Among the array of methodologies available, Welch’s method \cite{w1967} stands as a preferred technique due to its robustness in estimating the power spectral density of EEG signals.

In this investigation, attention is directed towards specific EEG frequency bands, namely Theta (4-8 Hz), Alpha (8-12 Hz), Beta (13-30 Hz), and Gamma ($>$30 Hz). These bands are conventionally associated with distinct neural activities and cognitive states, rendering them essential for analyzing engagement and attention levels.

Following the acquisition of EEG data, a preliminary step involves segmenting the data into sentence-specific intervals. This segmentation ensures that subsequent analyses are contextually aligned with the auditory stimuli presented. The power spectral density (PSD) \cite{psd} is computed with the segmented data. The PSD serves as a metric in neurophysiological signal processing, elucidating the power distribution across various frequency components of the EEG signal.

% To augment the depth of our analysis, the short-time Fourier transform (STFT) \cite{stft} was employed. Contrary to the global frequency perspective offered by the traditional Fourier transform, the STFT provides a time-localized frequency representation. This characteristic of the STFT is instrumental in capturing temporal variations in frequency components, a feature paramount in EEG analysis when evaluating transient cognitive processes.

%A widely utilized spectral density estimation technique in EEG feature extraction is through Welch’s method \cite{w1967}. EEG frequency bands, including Theta (4-8 Hz), Alpha (8-12 Hz), Beta (13-30 Hz), and Gamma ($>$30 Hz), were analyzed to assess the subject's engagement and attention levels. After we filtered EEG into sentence segments, we calculated the power spectral density (PSD) \cite{psd} based on this method.  PSD specifies the power levels of the frequency components present in a signal. 

%The EEG data were transformed into the time-frequency distribution using the short-time Fourier transform (STFT) \cite{stft}. STFT provides the time-localized frequency information for situations in which frequency components of a signal vary over time.

\subsubsection{NLP Features}
A commonly used technique to perform syntactic analysis on the sentence-level structure is producing a syntax tree \cite{syntaxtree}. This tree-based representation, often referred to as a syntactic tree or parse tree, provides a hierarchical structure that delineates the grammatical relationships and dependencies within phrases and sentences of human languages.

\begin{figure}
    \centering
    \includegraphics[width=1 \columnwidth]{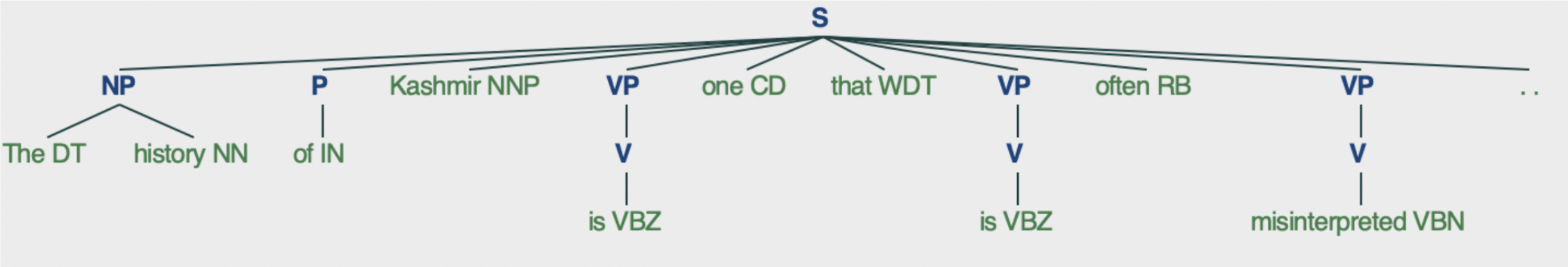}
    \caption{Syntax tree of a sentence from the lecture.}
    \label{fig:syntaxtree}
\end{figure}

The representation serves as a quantifiable metric for assessing the complexity of linguistic constructs. One such metric is the computation of the number of subtrees within the overarching syntax tree. Each subtree can be conceptualized as a node that governs its smaller tree structure, encapsulating specific syntactic relationships within a sentence segment. This hierarchical decomposition is illustrated in Figure~\ref{fig:syntaxtree}.

The derived number of subtrees is juxtaposed with participants' self-rated confusion levels, which provides a quantitative measure to assess the correlation between sentence complexity and perceived confusion. The EEG signals, which offer a biological perspective on cognitive processing, are also integrated into the analysis. A comprehensive understanding of the cognitive processing of linguistic structures is sought by examining the interplay between syntactic complexity, self-rated confusion, and EEG patterns.

Dependency parsing is utilized to analyze the relationships between words in sentences, allowing for the computation of the total dependency length \cite{dependencyparsing}. 
This technique facilitates the extraction of syntactic relationships between words, thereby allowing for the computation of the total dependency length \cite{dependencyparsing}. Such a length provides a quantitative measure of the interconnectedness and structural complexity of the sentence.

However, it is imperative to consider the potential confounding effect of sentence length on the perceived complexity. To mitigate this, the computed dependency length is normalized by dividing it by the total number of words present in the sentence. This normalization process ensures that the derived metric is independent of the sheer length of the sentence, focusing instead on the inherent syntactic relationships.

\begin{figure}
    \centering
    \includegraphics[width=1 \columnwidth]{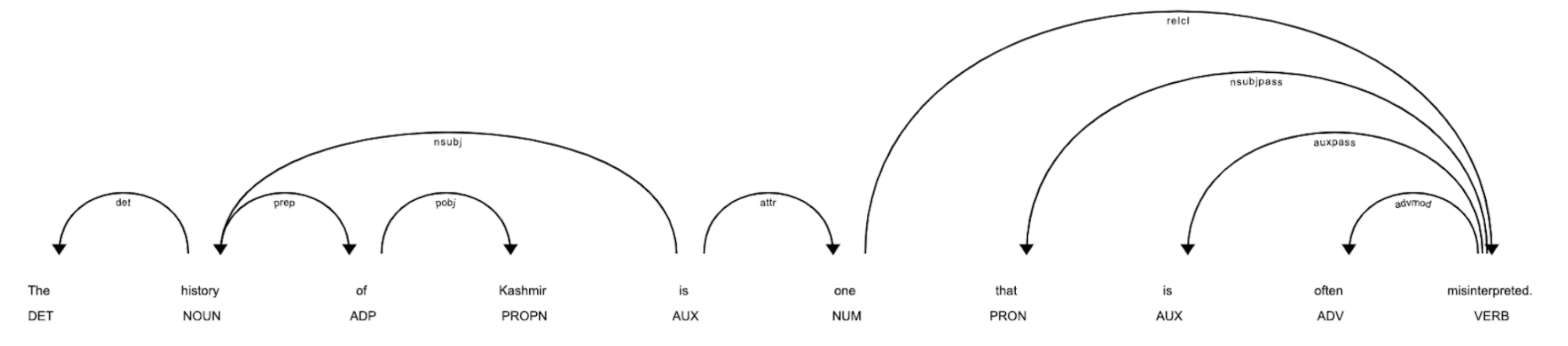}
    \caption{Example of a sentence from the lecture that has been dependency parsed.}
    \label{fig:parsing}
\end{figure}

To further elucidate these relationships, labels are incorporated within the dependency parsing framework. As depicted in Figure~\ref{fig:parsing}, these labels provide a visual representation of the nature and direction of dependencies between words, offering insights into the sentence's grammatical structure.

Beyond the total dependency length, other metrics are also derived to provide a more comprehensive understanding of sentence complexity. Specifically, the maximum distance between dependent words is computed, offering a measure of the farthest syntactic relationship within the sentence. Additionally, the average distances between all dependent word pairs are calculated, serving as an aggregate indicator of the overall complexity of the lecture content.

\subsection{Classification}

\color{black}After obtaining the aforementioned features, we concatenated the features together to generate a feature vector. By independently transforming the two modalities of information, we are able to separately or conjointly fit the baseline models. 

\color{black}In the proposed framework, we utilize an ensemble classification model in both predictive tasks -- predicting correctness and confusion.
This model ensemble multiple classification models, namely  XGBoost \cite{xgboost_paper}, Support Vector Machine \cite{svm_paper}, Logistic Regression \cite{lr_paper}, and Random Forest \cite{rf_paper}.
To validate the effectiveness of the proposed method, we employ ten-fold cross-validation. 
This method separates the participant data into ten segments, trains the model on nine segments, and tests it on the remaining segment.
This process repeats to ensure the data of all participants are served as testing data.
To achieve the best performance, we use GridSearch cross-validation technique to find a set of the most optimal hyper-parameter values \cite{gridsearch_paper}.
The overall classification framework is similar for both tasks; however, there are some detailed differences in the goal of these two tasks, which will be further discussed.

\subsubsection{\textcolor{black}{Predicting Correctness}}

Due to the characteristics of the Stanford Question Answering Dataset (SQuAD) we utilize, not every sentence is accompanied by a corresponding comprehension question. 
This inconsistency means that for some sentences, we lack the direct means to gauge comprehension. 
This gap in data poses a challenge in our experimental setup. 
To address and fill this gap, we employ the Label Spreading technique \cite{label_spreading}. 
This method predicts labels for data points that lack them by referencing other similar, labeled data points. 
During the training phase, we experiment with different data combinations as model input: solely EEG features, a combination of EEG and NLP features, and a set that contains features from EEG, NLP, and participants' self-rated confusion levels.
Finally, based on the input data, the model determines whether the participant would respond to the post-lecture assessment correctly.

\subsubsection{\textcolor{black}{Predicting Confusion}}
% 1. We need to explain the model, validation, and model concept.

For a predicting confusion task, our goal is to determine the level of confusion in students. 
To label the data, we use participants' subjective ratings, ranging from 1-10. 
Ratings above the median are classified as ``confused," while those below are labeled as ``not confused." 
We train separate models using two distinct data sources: one with only EEG features and the other with a combination of EEG and NLP features.
This approach allows us to assess the predictive power of each data type. 
A challenge we face is an imbalance in our dataset, where one class has more data than the other. 
To address this, we employ the SMOTE technique \cite{smote} to create synthetic data for the underrepresented class.
However, adding synthetic data can introduce noise or create closely neighboring points from opposite classes, which can obscure the decision boundary.
To address this, the Tomek Links method \cite{tomek} is applied.
This technique identifies and removes such closely neighboring points from the majority class, refining our dataset and enhancing the clarity of our model's decision-making.

\subsubsection{\textcolor{black}{Baseline Methods}}
In this study, we compare the performance of the proposed method against three commonly used baseline methods, namely, Logistic Regression (LR) \cite{lr_paper}, Random Forest (RF) \cite{rf_paper}, and Support Vector Machine (SVM) \cite{svm_paper} for predicting correctness in answering the post-lecture questions and confusion after listening to the lecture.
The input features and processing steps are the same across all the methods.
In this work, we use the open-source implementation of all the models provided by the Python scikit-learn library \cite{sklearn_paper}.

\section{Results}

In this work, two experiments were performed -- predicting confusion after listening to the lecture and predicting correctness in answering post-lecture questions.
The classification results are further discussed in this section.
% For predicting confusion, features from EEG and NLP were evaluated.
% For predicting correctness, features from EEG, EEG and NLP, and a combination of EEG, NLP, and confusion level, were evaluated.

In order to evaluate the performance of this work, accuracy (ACC) and F1-score (F1) are used.
Accuracy provides the ratio of all correctly predicted labels to all predicted labels, while F1 provides a harmonized measure between precision and recall, which is especially crucial in contexts with imbalanced datasets.

\begin{table}[t]
\centering
\caption{Classification performance for predicting students' correctness in answering the post-lecture question. (Results are reported with their standard deviation).}
\label{tab:result}
\begin{tabular}{ccccc}
\hline
\textbf{Method}                       & \textbf{Metric}         & \textbf{EEG} & \textbf{EEG+NLP}  & \textbf{EEG+NLP+CON}\\ \hline
\multirow{2}{*}{LR} & ACC  & $0.50\pm0.10$  & 0.56 ± 0.06   & 0.54 ± 0.09      \\ \cline{2-5} 
                                     & F1 & 0.56 ± 0.12  & 0.62 ± 0.07  &  0.61 ± 0.11     \\ \hline
\multirow{2}{*}{RF}       & ACC  & 0.54 ± 0.12  & 0.70 ± 0.11 & 0.65 ± 0.04          \\ \cline{2-5} 
                                     & F1 & 0.65 ± 0.11  & 0.77 ± 0.09 & 0.74 ± 0.05        \\ \hline
\multirow{2}{*}{SVM}                 & ACC  & 0.47 ± 0.09  & 0.58 ± 0.11 & 0.56 ± 0.10        \\ \cline{2-5} 
                                     & F1 & 0.56 ± 0.08  & 0.63 ± 0.10 & 0.64 ± 0.11        \\ \hline
\multirow{2}{*}{\textbf{Proposed}}      & ACC  & 0.54 ± 0.14  & 0.70 ± 0.08 & \textbf {0.70 ± 0.07}        \\ \cline{2-5} 
                                     & F1 & 0.65 ± 0.11  & 0.77 ± 0.10 & \textbf {0.78 ± 0.05}        \\ \hline
\end{tabular}
\begin{tablenotes}
  \small
  \item Note: ACC: Accuracy; CON: Self-rated Confusion Levels; F1: F1-score; LR: Logistic Regression; Proposed:~Proposed Classification Framework; RF:~Random Forest; SVM: Support Vector Machine.
\end{tablenotes}
\end{table}

\textcolor{black}{\subsection{Predicting Correctness}}
The classification results for predicting the correctness of students' responses to post-lecture true/false questions are shown in Table~\ref{tab:result}.
It can be observed that among the existing baseline models, the highest classification accuracy of $0.65\pm0.04$ and F1 of $0.74\pm0.05$ can be obtained when a combination of EEG, NLP, and self-rated confusion levels as input features and Random Forest algorithm used as a classifier.
However, the proposed classification framework (indicated as \textbf{Proposed} in the table) can outperform all existing baseline methods, yielding the best performance of $0.70\pm0.07$ in terms of accuracy and $0.78\pm0.05$ in terms of F1.

Table~\ref{tab:matrix} provides insights into the relationship between the self-rated confusion levels and the correctness of their answers.
The results show instances where participants reported being confused but still answered the question correctly and cases where participants claimed to understand but answered incorrectly.
This highlights the discrepancy between subjective perceptions of comprehension and objective measures of understanding.

\begin{table}[t]
\centering
\caption{Relation between subjective rating and correctness.}
\begin{tabular}{|l|c|c|c|}
\cline{1-4} & \textbf{Correct} & \textbf{Incorrect} & \textbf{TOTAL} \\ \cline{1-4}
\textbf{Confused} & 248 & 187 & 435 \\ \cline{1-4}
\textbf{Not Confused} & 206 & 47 & 253 \\ \cline{1-4}
\textbf{TOTAL} & 454 & 234 & 688 \\ \cline{1-4}
\end{tabular}
\label{tab:matrix}
\end{table}

% The proposed classification model utilized the semi-supervised algorithm, Label Spreading, to predict the accuracy of participants' responses to true/false statements. Various metrics, including EEG features only, combined EEG and NLP features, and participants' self-reported confusion rating scores, were employed in the analysis.
% The classification results, in terms of accuracy and F1 score, are shown in Table~\ref{tab:result}.

%$\begin{table}[t]
%\centering
%\caption{Classification performance for predicting students' correctness of answers.}
%\begin{tabular}{|c|ccc|}
%\cline{1-4}
%\multicolumn{1}{|l|}{} & \multicolumn{3}{c|}{\textbf{Features}} \\ \cline{1-4}
%\textbf{Metric}        & \multicolumn{1}{c|}{\textbf{EEG}} & \multicolumn{1}{c|}{\textbf{EEG+NLP}} & \textbf{EEG+NLP+Confusion } \\ \cline{1-4}
%\textit{accuracy$\pm$SD} & \multicolumn{1}{c|}{$0.72 \pm 0.09$} & \multicolumn{1}{c|}{$0.85 \pm 0.06$} & $0.80 \pm 0.08$  \\ \cline{1-4}
%\textit{f1\_score$\pm$SD} & \multicolumn{1}{c|}{$0.68 \pm 0.12$} & \multicolumn{1}{c|}{$0.83 \pm 0.07$} & $0.80\pm 0.05$ \\ \cline{1-4}
%\end{tabular}
%\label{tab:result}
%\end{table}

\textcolor{black}{\subsection{Predicting Confusion}}
Classification performance for predicting students' confusion level using EEG and the combination of EEG and NLP features are reported in Table \ref{tab:confusionresults}.
It can be observed that among all baseline models, Random Forest performs best when the combination of EEG and NLP are used as input features in which the best accuracy can be obtained of $0.65\pm0.07$.
With the proposed classification framework, although the reported accuracy of $0.61\pm0.07$ is slightly lower than the Random Forest method, the F1 has achieved $0.65\pm0.08$, which is the highest of all methods.

Compared to using only EEG features, incorporating NLP features demonstrated an improved performance for both accuracy and F1, indicating that a higher number of features helps the model learn better.

\begin{table}[t]
\centering
\caption{Classification performance for predicting students' confusion. (Results are reported with their standard deviation).}
\label{tab:confusionresults}
\begin{tabular}{cccc}
\hline
\textbf{Method}                       & \textbf{Metric}         & \textbf{EEG} & \textbf{EEG+NLP} \\ \hline
\multirow{2}{*}{LR} & ACC  & 0.54 ± 0.09  & 0.58 ± 0.07        \\ \cline{2-4} 
                                     & F1 & 0.52 ± 0.08  & 0.56 ± 0.01        \\ \hline
\multirow{2}{*}{RF}       & ACC  & 0.52 ± 0.08  & \textbf{0.65 ± 0.07}        \\ \cline{2-4} 
                                     & F1 & 0.47 ± 0.08  & 0.62 ± 0.05        \\ \hline
\multirow{2}{*}{SVM}                 & ACC  & 0.52 ± 0.10  & 0.60 ± 0.06        \\ \cline{2-4} 
                                     & F1 & 0.52 ± 0.10  & 0.57 ± 0.09        \\ \hline
\multirow{2}{*}{\textbf{Proposed}}      & ACC  & 0.49 ± 0.07  & 0.61 ± 0.07        \\ \cline{2-4} 
                                     & F1 & 0.54 ± 0.07  & \textbf{0.65 ± 0.08}       \\ \hline
\end{tabular}

\begin{tablenotes}
  \small
  \item Note: ACC: Accuracy; F1: F1-score; LR: Logistic Regression; Proposed:~Proposed Classification Framework; RF:~Random Forest; SVM: Support Vector Machine.
\end{tablenotes}
\end{table}

\section{Discussion}
%\textcolor{red}{
% Rewrite
%Although the results give a unique insight into combining linguistics and EEG analysis, the NLP features selected only highlight the general ideas of sentence complexity through dependency parsing and syntax trees.
%Interestingly, 248 instances were found where participants claimed not to understand the sentence but still answered correctly, suggesting guessing.
%Alternatively, it could indicate that the sentence was confusing, but the content itself was easy to grasp, allowing educated guesses.
%There were 47 instances where participants misunderstood the passage, highlighting the potential for misperception even when they believe they understand.
%This underscores the importance of predicting correct/incorrect answers to ensure genuine comprehension of delivered information.}

% \textcolor{red}{Knot will continue with discussion. The rest is done but proof-read and check grammar are needed.}

%\textcolor{red}{
%Discussion section should discuss the results and rationale behind.
%It should DISCUSS results but NOT repeat what was already stated in section Results.
%For example, 1. Discuss WHY our model performs better that state-of-the-art.
%2. Discuss why combination of features perform better than single feature.
%Discuss separately for predicting confusion and predicting correctness on each paragraph.
%This section is the hardest section to write.
%My advise for you is "To look at the results and come up with the question why this is so?"
%}

From the results, the proposed classification framework achieves the best performance of 0.70 in terms of accuracy and 0.78 in terms of F1 in predicting the correctness of students' post-lecture responses, surpassing the highest classification accuracy of 0.65 and F1 of 0.74 from the baseline model.

The proposed framework's proficiency in predicting the correctness of post-lecture responses is also rooted in this integration approach. By integrating features from EEG, NLP, and self-rated confusion levels, the proposed framework achieves a holistic understanding of a participant's comprehension. EEG data highlights cognitive engagement, while NLP features can reveal the depth of content processing. The inclusion of self-rated confusion levels adds an additional layer, offering insights into a student's perceived understanding. Together, these features ensure a rounded assessment, increasing the chances of accurately predicting whether a student's response aligns with the content's complexity and intent.

The enhanced performance of our framework in predicting confusion is attributed to the integration of EEG and NLP features. EEG captures the underlying neural activity, offering valuable insights into participants' brain states. However, to provide a comprehensive understanding, it's essential to contextualize these readings with the text's inherent complexity and participants' subjective experiences. This ensures that the observed cognitive processes align with the expected demands of the text. NLP syntactic analysis complements this by analyzing language patterns, detecting signs of confusion that might not be evident through EEG alone. The combination of these features provides a robust framework, ensuring that all potential indicators of confusion, be they from brain activity or linguistic patterns, are considered.

The decision to use an ensemble stacking method stemmed from the individual strengths and limitations of each model in our study. While Logistic Regression is adept at handling linear data, it can falter with non-linear complexities. 
Random Forest, though versatile with diverse datasets, might overfit on noisy data. 
SVM excels in high-dimensional spaces but can be cumbersome with vast datasets and demands precise parameter tuning. 
XGBoost, with its optimization prowess, still carries inherent biases.

Ensemble stacking addresses these limitations by amalgamating predictions. For example, non-linearities missed by Logistic Regression might be captured by Random Forest or XGBoost. The intricate decision boundaries identified by SVM can fill gaps where other models generalize. The ensemble harnesses collective strengths through stacking, often yielding predictions superior to any individual model.

Data preprocessing techniques are also crucial in refining performance of the proposed framework. SMOTE addresses class imbalance by generating synthetic samples, ensuring unbiased model training. 
Tomek Links further refines this by removing closely paired instances from opposite classes, enhancing the decision boundary. 
Label spreading, a semi-supervised technique, utilizes both labeled and unlabeled data to boost prediction accuracy, especially when labeled data is limited. 
In conjunction with stacking, these techniques optimize the proposed framework for predicting student comprehension and response accuracy.

%\textcolor{red}{
%Furthermore, an intriguing observation emerged regarding the potential disconnect between students' subjective comprehension ratings and their actual performance in response accuracy. This disparity underscores the complexities inherent in gauging comprehension. While subjective ratings offer a reflection of students' perceived understanding, they may not always align with objective measures of comprehension. This observation accentuates the challenges researchers face in obtaining precise, reliable measurements of comprehension and highlights the need for multi-faceted assessment tools in educational research.
%}

Although the performance of the proposed framework exceeds the baselines, they have certain limitations when it comes to improving the lecturing process comprehensively.
The experiment only captures understanding at the sentence level, which is the initial step toward holistically enhancing the process.
Moreover, the absence of teacher-student interaction in our experiment limits its ecological validity, as the participants were solely exposed to audio files.

% Future work could benefit from using NLP text summarization for qualitative assessment of student understanding, surpassing the limitations of narrow multiple-choice comprehension tests. Additionally, exploring brain synchronization as a biometric for information transfer between teachers and students using EEGs could offer valuable insights. Simulating a real classroom environment with an actual teacher would enhance the study's applicability. 

\section{Conclusion}
This study introduces an ensemble classification framework aimed at predicting students' confusion levels and the accuracy of post-lecture responses by leveraging integrated EEG and NLP features. 
To enhance classification performance, self-rated confusion levels are incorporated as additional features. 
The proposed approach achieves a remarkable accuracy of 0.70 and an F1 score of 0.78, surpassing the baseline models' highest accuracy of 0.65 and F1 score of 0.74. 
These findings underscore the potential of utilizing EEG and NLP features alongside subjective data and the ensemble classification method for effectively predicting students' lecture comprehension.

\textcolor{red}{
 %This study introduces an innovative methodology that combines EEG biomarkers with linguistic analytics provided by NLP features. This integration aims to assess student comprehension and pinpoint instances of confusion subsequent to their engagement with classroom lectures.
 }

%\textcolor{red}{
%Our experimental protocol, distinct in its design, facilitates real-time predictive analysis post each sentence, offering a precise metric of comprehension. Notably, the achieved accuracies stand at 0.70 and 0.61 for both predicting correctness and confusion, respectively. These metrics underscore the efficacy of fusing subjective self-assessments with objective performance measures.
%\textcolor{red}{
%Furthermore, the commendable performance of our model can be attributed to the holistic assessment approach, made feasible by the confluence of EEG and NLP features. It's noteworthy to mention that, in comparison to the state-of-the-art, which achieved an accuracy of 75\% \cite{mooc} utilizing solely EEG data, our integrated approach demonstrates a significant advancement in the field.
%}


\section*{Acknowledgment}
We would like to acknowledge the support received from the National Science and Technology Development Agency (NSTDA), Vidyasirimedhi Institute of Science and Technology (VISTEC), and Ruamrudee International School (RIS) throughout this work.



%\bibliographystyle{IEEEtran}
% \bibliographystyle{IEEEtran}
% \bibliography{Reference}
\printbibliography

\end{document}